\documentclass{article} 
\usepackage{iclr2018_conference,times}
\usepackage{hyperref}
\usepackage{url}
\usepackage{graphicx}
\usepackage{amssymb}
\usepackage{amsmath}
\usepackage{multirow}

\title{Pixel Deconvolutional Networks}

\author{
Hongyang Gao\\
Washington State University\\
\texttt{hongyang.gao@wsu.edu}\\
\And
Hao Yuan\\
Washington State University\\
\texttt{hao.yuan@wsu.edu}\\
\And
Zhengyang Wang\\
Washington State University\\
\texttt{zwang6@eecs.wsu.edu}\\
\AND
Shuiwang Ji\\
Washington State University\\
\texttt{sji@eecs.wsu.edu}\\
}

\iclrfinalcopy 

\begin{document}

\maketitle

\begin{abstract}
Deconvolutional layers have been widely used in a variety of deep
models for up-sampling, including encoder-decoder networks for
semantic segmentation and deep generative models for unsupervised
learning. One of the key limitations of deconvolutional operations
is that they result in the so-called checkerboard problem. This is
caused by the fact that no direct relationship exists among adjacent
pixels on the output feature map. To address this problem, we
propose the pixel deconvolutional layer (PixelDCL) to establish
direct relationships among adjacent pixels on the up-sampled feature
map. Our method is based on a fresh interpretation of the regular
deconvolution operation. The resulting PixelDCL can be used to
replace any deconvolutional layer in a plug-and-play manner without
compromising the fully trainable capabilities of original models.
The proposed PixelDCL may result in slight decrease in efficiency,
but this can be overcome by an implementation trick. Experimental
results on semantic segmentation demonstrate that PixelDCL can
consider spatial features such as edges and shapes and yields more
accurate segmentation outputs than deconvolutional layers. When used
in image generation tasks, our PixelDCL can largely overcome the
checkerboard problem suffered by regular deconvolution operations.
\end{abstract}

\section{Introduction}\label{introduction}
Deep learning methods have shown great promise in a variety of
artificial intelligence tasks such as image
classification~\citep{Alex12,vgg14}, semantic
segmentation~\citep{deconv15,shelhamer2016fully,ronneberger2015u},
and natural image
generation~\citep{goodfellow2014generative,kingma2014stochastic,oord2016pixel}.
Some key network layers, such as convolutional
layers~\citep{lecun1998gradient}, pooling layers, fully connected
layers and deconvolutional layers, have been frequently used to
create deep models for different tasks. Deconvolutional layers, also
known as transposed convolutional
layers~\citep{vedaldi2015matconvnet}, are initially proposed
in~\citep{zeiler2010deconvolutional,zeiler2011adaptive}. They have
been primarily used in deep models that require up-sampling of
feature maps, such as generative
models~\citep{radford2015unsupervised,makhzani2015winner,rezende2014stochastic}
and encoder-decoder architectures~\citep{ronneberger2015u,deconv15}.
Although deconvolutional layers are capable of producing larger
feature maps from smaller ones, they suffer from the problem of
checkerboard artifacts~\citep{odena2016deconvolution}. This greatly
limits deep model's capabilities in generating photo-realistic
images and producing smooth outputs on semantic segmentation. To
date, very little efforts have been devoted to improving the
deconvolution operation.

\begin{figure}[t]
  \begin{center}
    \includegraphics[width=0.9\textwidth,height=6cm]{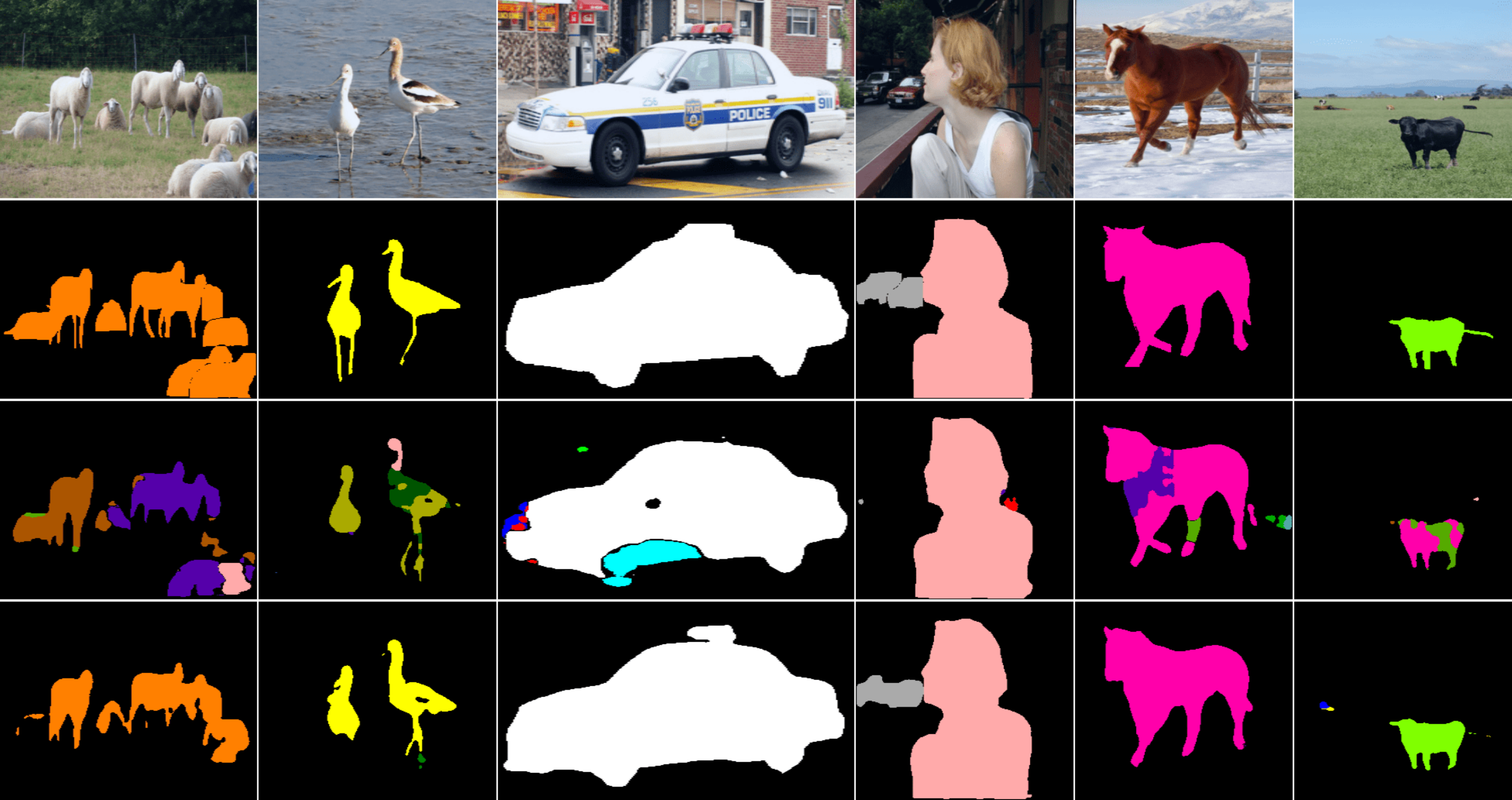}
    \caption{
      Comparison of semantic segmentation results. The first and second rows
      are images and ground true labels, respectively. The third and fourth
      rows are the results of using regular deconvolution and our proposed
      pixel deconvolution PixelDCL, respectively.
    }
    \label{fig:pre_pascal_result}
  \end{center}
\end{figure}

In this work, we propose a simple, efficient, yet effective method, known as
the pixel deconvolutional layer (PixelDCL), to address the checkerboard
problem suffered by deconvolution operations. Our method is motivated from a
fresh interpretation of deconvolution operations, which clearly pinpoints the
root of checkerboard artifacts. That is, the up-sampled feature map generated
by deconvolution can be considered as the result of periodical shuffling of
multiple intermediate feature maps computed from the input feature map by
independent convolutions. As a result, adjacent pixels on the output feature
map are not directly related, leading to the checkerboard artifacts. To
overcome this problem, we propose the pixel deconvolutional operation to be
used in PixelDCL. In this new layer, the intermediate feature maps are
generated sequentially so that feature maps generated in a later stage are
required to depend on previously generated ones. In this way, direct
relationships among adjacent pixels on the output feature map have been
established. Sequential generation of intermediate feature maps in PixelDCL
may result in slight decrease in computational efficiency, but we show that
this can be largely overcome by an implementation trick. Experimental results
on semantic segmentation (samples in Figure~\ref{fig:pre_pascal_result}) and
image generation tasks demonstrate that the proposed PixelDCL can effectively
overcome the checkerboard problem and improve predictive and generative
performance.

Our work is related to the pixel recurrent neural networks
(PixelRNNs)~\citep{oord2016pixel} and
PixelCNNs~\citep{van2016conditional,reed2017parallel}, which are generative
models that consider the relationship among units on the same feature map.
They belong to a more general class of autoregressive methods for probability
density estimation~\citep{germain2015made,gregor2015draw,larochelle2011neural}.
By using masked convolutions in training, the training time of PixelRNNs and
PixelCNNs is comparable to that of other generative models such as generative
adversarial networks (GANs)~\citep{goodfellow2014generative,reed2016generative}
and variational auto-encoders
(VAEs)~\citep{kingma2014stochastic,johnson2016composing}. However, the prediction
time of PixelRNNs or PixelCNNs is very slow since it has to generate images
pixel by pixel. In contrast, our PixelDCL can be used to replace any
deconvolutional layer in a plug-and-play manner, and the slight decrease in
efficiency can be largely overcome by an implementation trick.


\section{Pixel Deconvolutional Layers and Networks}\label{sec:pixel_deconv}

We introduce deconvolutional layers and analyze the cause of
checkerboard artifacts in this section. We then propose the pixel
deconvolutional layers and the implementation trick to improve
efficiency.

\subsection{Deconvolutional Layers}\label{sec:deconv}

Deconvolutional networks and deconvolutional layers are proposed
in~\citep{zeiler2010deconvolutional,zeiler2011adaptive}. They have
been widely used in deep models for applications such as semantic
segmentation~\citep{deconv15} and generative
models~\citep{kingma2014stochastic,goodfellow2014generative,oord2016pixel}.
Many encoder-decoder architectures use deconvolutional layers in
decoders for up-sampling. One way of understanding deconvolutional
operations is that the up-sampled output feature map is obtained by
periodical shuffling of multiple intermediate feature maps obtained
by applying multiple convolutional operations on the input feature
maps~\citep{shi2016real}.

\begin{figure}[ht]
  \centering
  \includegraphics[width=\textwidth]{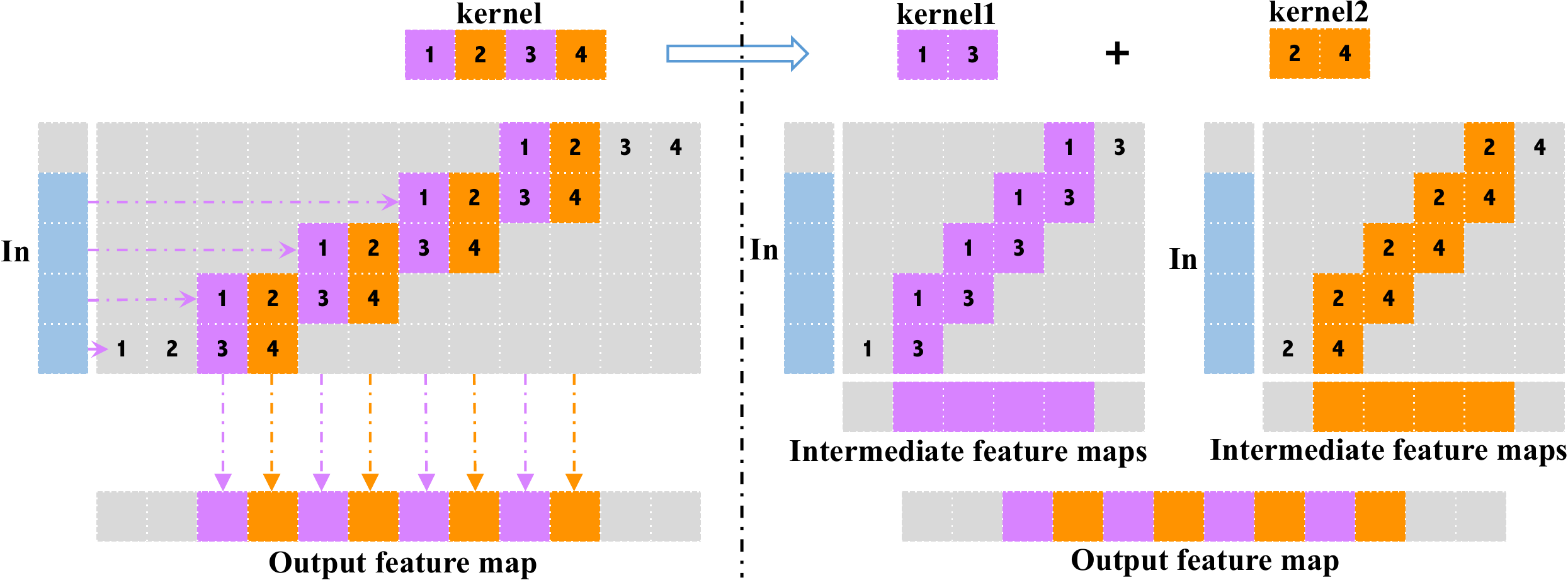}
    \caption{
    Illustration of 1D deconvolutional operation. In this deconvolutional
    layer, a 4$\times$1 feature map is up-sampled to an 8$\times$1 feature
    map. The left figure shows that each input unit passes through an 1$\times$4
    kernel. The output feature map is obtained as the sum of values
    in each column. It can be seen from this figure that the purple outputs are only related
    to (1, 3) entries in the kernel, while the orange outputs are only related to (2, 4) entries in the kernel.
    Therefore, 1D deconvolution can be decomposed as
    two convolutional operations shown in the right figure. The two intermediate
    feature maps generated by convolutional operations are dilated and combined
    to obtain the final output. This indicates that the standard deconvolutional operation
    can be decomposed into multiple convolutional operations.
  }
  \label{fig:deconv1d}
\end{figure}

\begin{figure}[t]
  \centering
  \includegraphics[width=\textwidth]{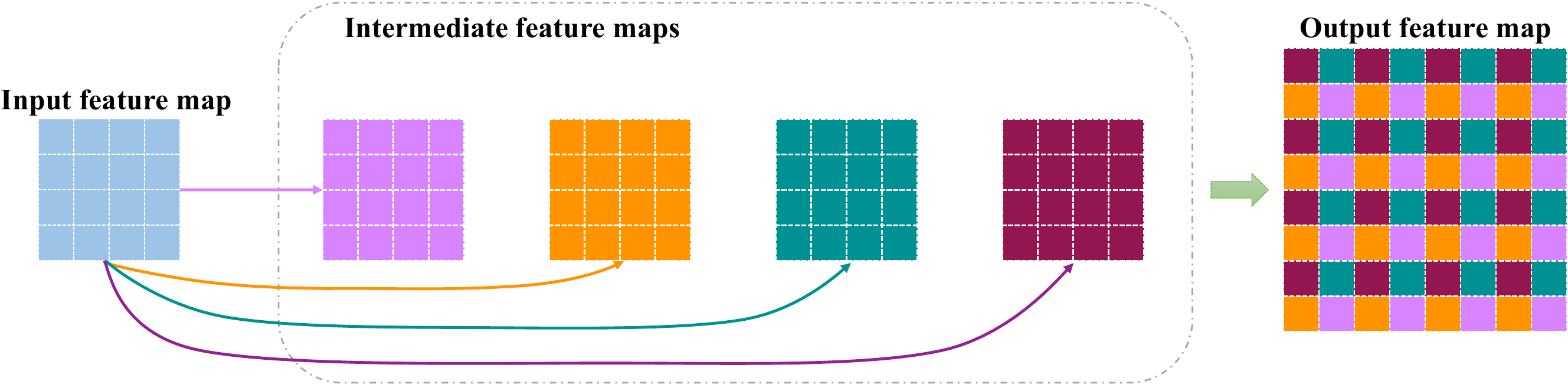}
    \caption{
    Illustration of 2D deconvolutional operation. In this deconvolutional
    layer, a 4$\times$4 feature map is up-sampled to an 8$\times$8 feature
    map. Four intermediate feature maps (purple, orange, blue, and red)
    are generated using four different convolutional kernels. Then these
    four intermediate feature maps are shuffled and combined to produce the
    final 8$\times$8 feature map. Note that the four intermediate feature maps
    rely on the input feature map but with no direct relationship among
    them.
  }
  \label{fig:deconv}
\end{figure}

This interpretation of deconvolution in 1D and 2D is illustrated in
Figures~\ref{fig:deconv1d} and~\ref{fig:deconv}, respectively. It is
clear from these illustrations that standard deconvolutional
operation can be decomposed into several convolutional operations
depending on the up-sampling factor. In the following, we assume the
up-sampling factor is two, though deconvolution operations can be
applied to more generic settings. Formally, given an input feature
map $F_{in}$, a deconvolutional layer can be used to generate an
up-sampled output $F_{out}$ as follows:
\begin{equation}
\begin{aligned}
  F_{1} = F_{in} \circledast k_1,  \qquad F_{2} &= F_{in} \circledast k_2, \qquad
  F_{3} = F_{in} \circledast k_3, \qquad F_{4} = F_{in} \circledast k_4, \\
  F_{out} &= F_{1} \oplus F_{2} \oplus F_{3} \oplus F_{4},
\end{aligned}\label{eq:deconv}
\end{equation}
where $\circledast$ denotes the convolutional operation and $\oplus$
denotes the periodical shuffling and combination operation as in
Figure~\ref{fig:deconv}, $F_{i}$ is the intermediate feature map
generated by the corresponding convolutional kernel $k_i$ for
$i=1,\cdots,4$.

\begin{figure}[t]
  \begin{center}
    \includegraphics[width=\textwidth]{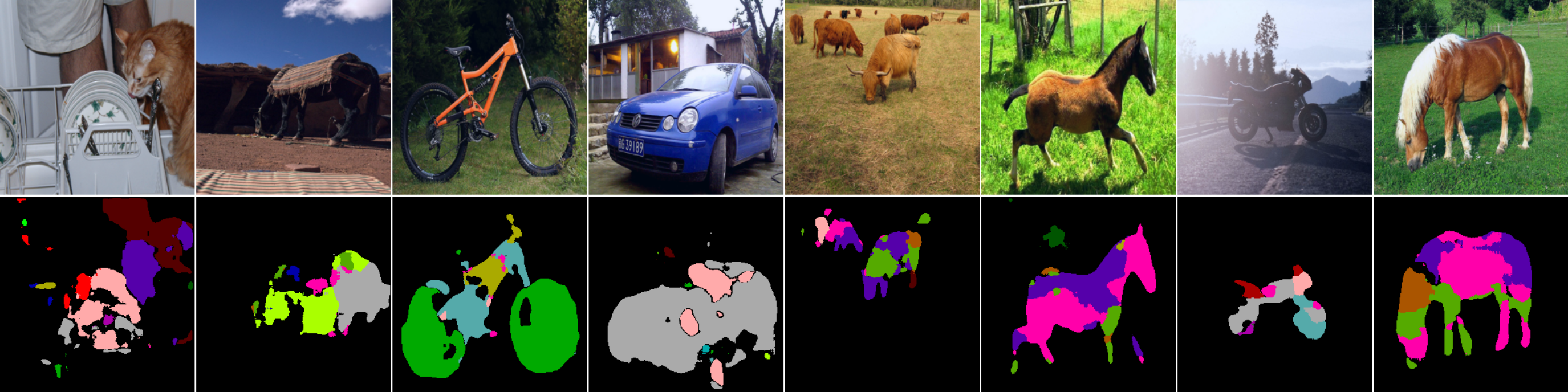}
    \caption{
      Illustration of the checkerboard problem in semantic segmentation
      using deconvolutional layers. The first and second rows
      are the original images and semantic segmentation results, respectively.
    }
    \label{fig:bad_result}
  \end{center}
\end{figure}

It is clear from the above interpretation of deconvolution that
there is no direct relationship among these intermediate feature
maps since they are generated by independent convolutional kernels.
Although pixels of the same position on intermediate feature maps
depend on the same receptive field of the input feature map, they
are not directly related to each other. Due to the periodical
shuffling operation, adjacent pixels on the output feature map are
from different intermediate feature maps. This implies that the
values of adjacent pixels can be significantly different from each
other, resulting in the problem of checkerboard
artifacts~\citep{odena2016deconvolution} as illustrated in Figure~\ref{fig:bad_result}.
One way to alleviate
checkerboard artifacts is to apply post-processing such as
smoothing~\citep{li2001simple}, but this adds additional
complexity to the network and makes the entire network not fully
trainable. In this work, we propose the pixel deconvolutional
operation to add direct dependencies among intermediate feature
maps, thereby making the values of adjacent pixels close to each
other and effectively solving the checkerboard artifact problem. In
addition, our pixel deconvolutional layers can be easily used to
replace any deconvolutional layers without compromising the fully
trainable capability.

\subsection{Pixel Deconvolutional Layers}\label{sec:pixelDeconv}

\begin{figure}[t]
  \centering
  \includegraphics[width=\textwidth]{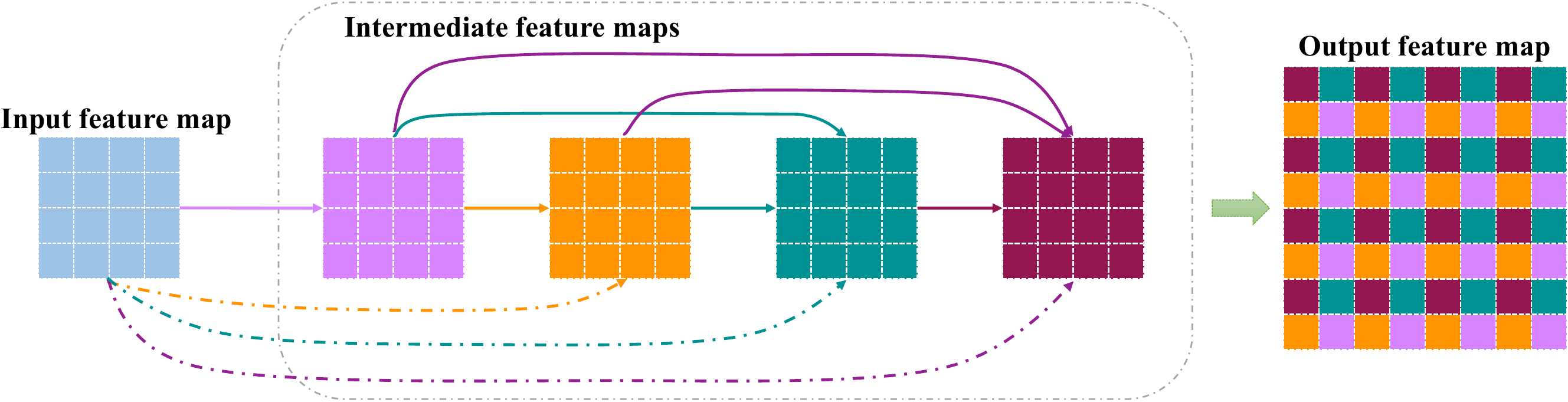}
  \caption{
    Illustration of iPixelDCL and PixelDCL described in
    section~\ref{sec:pixelDeconv}. In iPixelDCL, there are additional
    dependencies among intermediate feature maps. Specifically, the four
    intermediate feature maps are generated sequentially. The purple feature
    map is generated from the input feature map (blue). The orange feature map
    is conditioned on both the input feature map and the purple feature map
    that has been generated previously. In this way, the green feature map
    relies on the input feature map, purple and orange intermediate feature
    maps. The red feature map is generated based on the input feature map,
    purple, orange, and green intermediate feature maps. We also propose to
    move one step further and allow only the first intermediate feature map to
    depend on the input feature map. This gives rise to PixelDCL. That is, the
    connections indicated by dashed lines are removed to avoid repeated
    influence of the input feature map. In this way, only the first feature map
    is generated from the input and other feature maps do not directly rely on
    the input. In PixelDCL, the orange feature map only depends on the purple
    feature map. The green feature map relies on the purple and orange feature
    maps. The red feature map is conditioned on the purple, orange, and green
    feature maps. The information of the input feature map is delivered to
    other intermediate feature maps through the first intermediate feature map
    (purple).
  }
  \label{fig:method}
\end{figure}

To solve the checkerboard problem in deconvolutional layers, we
propose the pixel deconvolutional layers (PixelDCL) that can add
dependencies among intermediate feature maps. As adjacent pixels are
from different intermediate feature maps, PixelDCL can build direct
relationships among them, thus solving the checkerboard problem. In
this method, intermediate feature maps are generated sequentially
instead of simultaneously. The intermediate feature maps generated
in a later stage are required to depend on previously generated
ones. The primary purpose of sequential generation is to add
dependencies among intermediate feature maps and thus adjacent
pixels in final output feature maps. Finally, these intermediate
feature maps are shuffled and combined to produce final output
feature maps. Compared to Eqn.~\ref{eq:deconv}, $F_{out}$ is
obtained as follows:
\begin{equation}
\begin{aligned}
  F_{1} &= F_{in} \circledast k_1, \qquad
  & F_{2} &= [F_{in}, F_{1}] \circledast k_2, \\
  F_{3} &= [F_{in}, F_{1}, F_{2}] \circledast k_3, \qquad
  & F_{4} &= [F_{in}, F_{1}, F_{2}, F_{3}] \circledast k_4, \\
  F_{out} &= F_{1} \oplus F_{2} \oplus F_{3} \oplus F_{4},&
\end{aligned}\label{eq:con_pixel_deconv}
\end{equation}
where $[\cdot, \cdot]$ denotes the juxtaposition of feature maps.
Note that in Eqn.~\ref{eq:con_pixel_deconv}, $k_i$ denotes a set of
kernels as it involves convolution with the juxtaposition of
multiple feature maps. Since the intermediate feature maps in
Eqn.~\ref{eq:con_pixel_deconv} depend on both the input feature map
and the previously generated ones, we term it input pixel
deconvolutional layer (iPixelDCL). Through this process, pixels on
output feature maps will be conditioned not only on input feature
maps but also on adjacent pixels. Since there are direct
relationships among intermediate feature maps and adjacent pixels,
iPixelDCL is expected to solve the checkerboard problem to some
extent. Note that the relationships among intermediate feature maps
can be very flexible. The intermediate feature maps generated later
on can rely on part or all of previously generated intermediate
feature maps. This depends on the design of pixel dependencies in
final output feature maps. Figure~\ref{fig:method} illustrates a
specific design of sequential dependencies among intermediate
feature maps.

In iPixelDCL, we add dependencies among generated intermediate
feature maps, thereby making adjacent pixels on final output feature
maps directly related to each other. In this process, the
information of the input feature map is repeatedly used when generating
intermediate feature maps. When generating the intermediate feature
maps, information from both the input feature map and previous
intermediate feature maps is used. Since previous intermediate
feature maps already contain information of the input feature map, the
dependencies on the input feature map can be removed. Removing
such dependencies for some intermediate feature
maps can not only improve the computational efficiency but also
reduce the number of trainable parameters in deep models.

In this simplified pixel deconvolutional layer, only the first
intermediate feature map will depend on the input feature map. The
intermediate feature maps generated afterwards will only depend on
previously generated intermediate feature maps. This will simplify
the dependencies among pixels on final output feature map. In this
work, we use PixelDCL to denote this simplified design. Our
experimental results show that PixelDCL yields better performance
than iPixelDCL and regular deconvolution. Compared to
Eqn.~\ref{eq:con_pixel_deconv}, $F_{out}$ in PixelDCL is obtained as
follows:
\begin{equation}
\begin{aligned}
  F_{1} &= F_{in} \circledast k_1, \qquad
  & F_{2} &= F_{1} \circledast k_2, \\
  F_{3} &= [F_{1}, F_{2}] \circledast k_3, \qquad
  & F_{4} &= [F_{1}, F_{2}, F_{3}] \circledast k_4, \\
  F_{out} &= F_{1} \oplus F_{2} \oplus F_{3} \oplus F_{4}.
\end{aligned}\label{eq:pixel_deconv}
\end{equation}

PixelDCL is illustrated in Figure~\ref{fig:method} by removing the
connections denoted with dash lines. When analyzing the
relationships of pixels on output feature maps, it is clear that
each pixel will still rely on adjacent pixels. Therefore, the
checkerboard problem can be solved with even better computational
efficiency. Meanwhile, our experimental results demonstrate that the
performance of models with these simplified dependencies is even
better than that with complete connections. This demonstrates that
repeated dependencies on the input may not be necessary.

\subsection{Pixel Deconvolutional Networks}\label{sec:pixelNet}

\begin{figure}[ht]
  \centering
  \includegraphics[width=\textwidth]{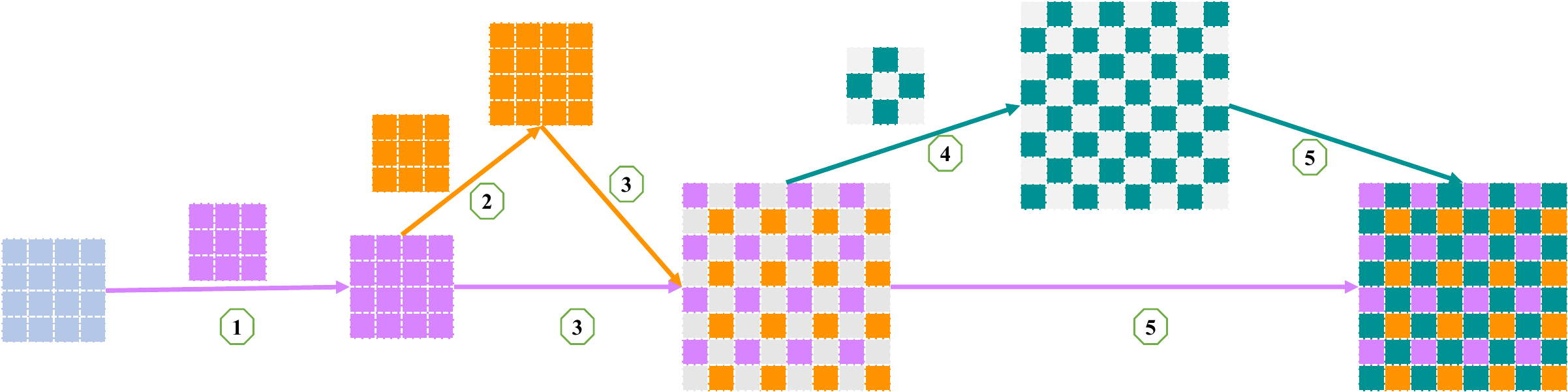}
  \caption{
    An efficient implementation of the pixel deconvolutional layer. In this
    layer, a 4$\times$4 feature map is up-sampled to a 8$\times$8 feature map.
    The purple feature map is generated through a 3$\times$3
    convolutional operation from the input feature map (step 1). After that,
    another 3$\times$3 convolutional operation is applied on the purple
    feature map to produce the orange feature map (step 2). The purple and
    orange feature maps are dilated and added together to form a larger
    feature map (step 3). Since there is no relationship between the last two
    intermediate feature maps, we can apply a masked 3$\times$3 convolutional
    operation, instead of two separate 3$\times$3 convolutional operations
    (step 4). Finally, the two large feature maps are combined to generate the
    final output feature map (step 5).
  }
  \label{fig:imple}
\end{figure}

Pixel deconvolutional layers can be applied to replace any
deconvolutional layers in various models involving up-sampling
operations such as U-Net~\citep{ronneberger2015u},
VAEs~\citep{kingma2014stochastic} and GANs~\citep{goodfellow2014generative}.
By replacing deconvolutional layers with pixel deconvolutional
layers, deconvolutional networks become pixel deconvolutional
networks (PixelDCN). In U-Net for semantic segmentation, pixel
deconvolutional layers can be used to up-sample from low-resolution
feature maps to high-resolution ones. In VAEs, they can be applied
in decoders for image reconstruction. The generator networks in GANs
typically use deep model~\citep{radford2015unsupervised} and thus can
employ pixel deconvolutional layers to generate large images. In our
experiments, we evaluate pixel deconvolutional layers in U-Net and
VAEs. The results show that the performance of pixel deconvolutional
layers outperforms deconvolutional layers in these networks.

In practice, the most frequently used up-sampling operation is to
increase the height and width of input feature maps by a factor of
two, e.g., from 2$\times$2 to 4$\times$4. In this case, the pixels
on output feature maps can be divided into four groups as in
Eqn.~\ref{eq:deconv}. The dependencies can be defined as in
Figure~\ref{fig:method}. When implementing pixel deconvolutional
layers, we design a simplified version to reduce sequential
dependencies for better parallel computation and training efficiency
as illustrated in Figure~\ref{fig:imple}.

In this design, there are four intermediate feature maps. The first
intermediate feature map depends on the input feature map. The
second intermediate feature map relies on the first intermediate
feature map. The third and fourth intermediate feature maps are
based on both the first and the second feature maps. Such simplified
relationships enable the parallel computation for the third and
fourth intermediate feature maps, since there is no dependency
between them. In addition, the masked convolutional operation can be
used to generate the last two intermediate feature maps. As has been
mentioned already, a variety of different dependencies relations can
be imposed on the intermediate feature maps. Our simplified design
achieves reasonable balance between efficiency and performance.
Our code is publicly
available\footnote{https://github.com/divelab/PixelDCN}



\section{Experimental Studies}\label{experiments}

In this section, we evaluate the proposed pixel deconvolutional
methods on semantic segmentation and image generation tasks in
comparison to the regular deconvolution method. Results show that
the use of the new pixel deconvolutional layers improves performance
consistently in both supervised and unsupervised learning settings.


\subsection{Semantic Segmentation}

\begin{figure}[t]
  \centering
  \includegraphics[width=\textwidth,height=5cm]{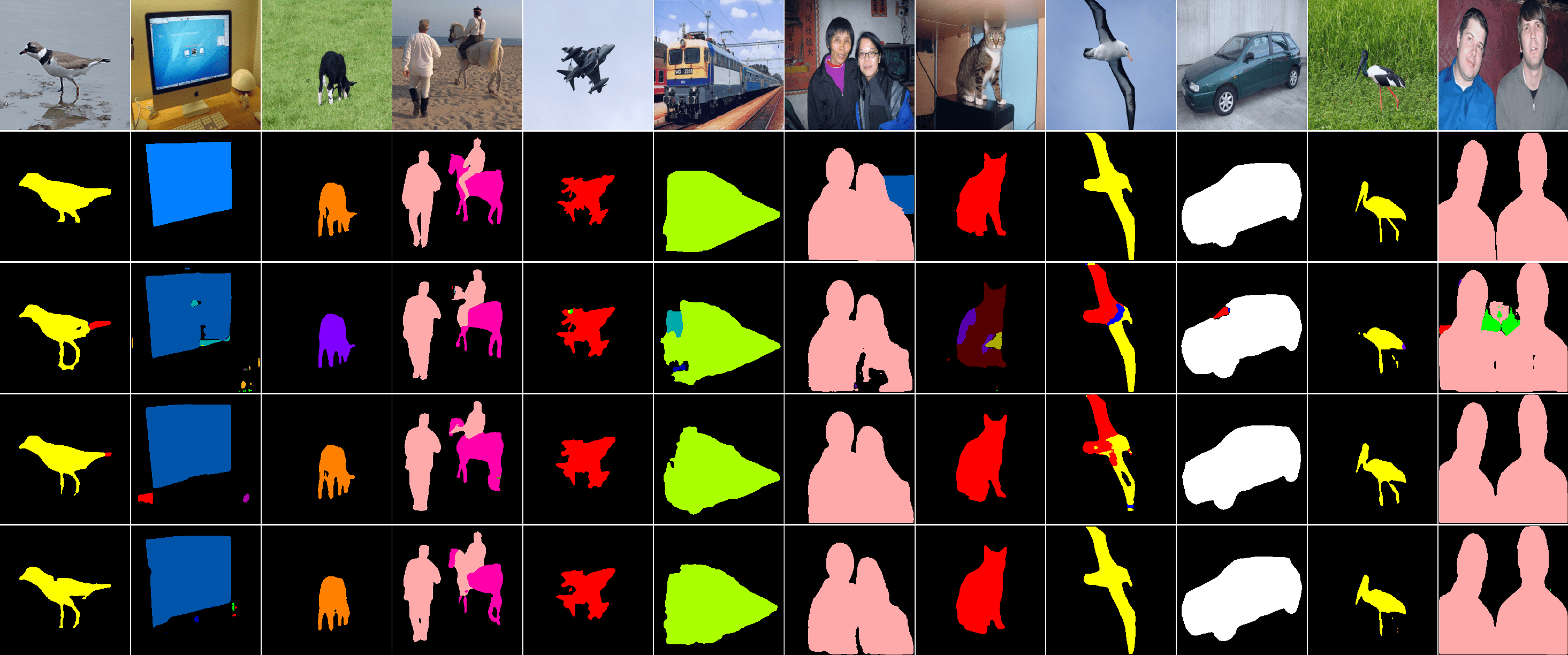}
  \caption{
    Sample segmentation results on the PASCAL 2012 segmentation dataset using
    training from scratch models. The
    first and second rows are the original images and the corresponding ground
    truth, respectively. The third, fourth, and fifth rows are the
    segmentation results of models using deconvolutional layers, iPixelDCL,
    and PixelDCL, respectively.
  }
  \label{fig:pascal_result}
\end{figure}

\textbf{Experimental Setup:} We use the PASCAL 2012 segmentation
dataset~\citep{everingham2010pascal} and MSCOCO 2015 detection
dataset~\citep{lin2014microsoft} to evaluate the proposed pixel
deconvolutional methods in semantic segmentation tasks. For both
datasets, the images are resized to 256$\times$256$\times$3 for
batch training. Our models directly predict the label for each pixel
without any post-processing. Here we examine our models in two ways:
training from scratch and fine-tuning from state-of-art
model such as DeepLab-ResNet.

For the training from scratch experiments, we use the U-Net
architecture~\citep{ronneberger2015u} as our base model as it has
been successfully applied in various image segmentation tasks. The
network consists of four blocks in the encoder path and four
corresponding blocks in the decoder path. Within each decoder block,
there is a deconvolutional layer followed by two convolutional
layers. The final output layer is adjusted based on the number of
classes in the dataset. The PASCAL 2012 segmentation dataset has 21
classes while the MSCOCO 2015 detection dataset has 81 classes. As
the MSCOCO 2015 detection dataset has more classes than the PASCAL
2012 segmentation dataset, the number of feature maps in each layer
for this dataset is doubled to accommodate more output channels. The
baseline U-Net model employs deconvolutional layers within the
decoder path to up-sample the feature maps. We replace the
deconvolutional layers with our proposed pixel deconvolutional
layers (iPixelDCL) and their simplified version (PixelDCL) while
keeping all other variables unchanged. The kernel size in DCL is
6$\times$6, which has the same number of parameters as iPixelDCL
with 4 sets of 3$\times$3 kernels, and more parameters than PixelDCL
with 2 sets of 3$\times$3 and 1 set of 2$\times$2 kernels. This will
enable us to evaluate the new pixel deconvolutional layers against
the regular deconvolutional layers while controlling all other
factors.

For the fine-tuning experiments, we fine-tune our models based on
the architecture of DeepLab-ResNet~\citep{CP2016DeepLab}. The
DeepLab-ResNet model is fine-tuned from ResNet101~\citep{he2016deep}
and also use external data for training. The strategy of using
external training data and fine-tuning from classic ResNet101
greatly boosts the performance of the model on both accuracy and
mean IOU. The output of DeepLab-ResNet is eight times smaller than
the input image on the height and width dimensions. In order to
recover the original dimensions, we add three up-sampling blocks,
each of which up-samples the feature maps by a factor of 2. For each
up-sampling block, there is a deconvolutional layer followed by a
convolutional layer. By employing the same strategy, we replace the
deconvolutional layer by PixelDCL and iPixelDCL using kernels of the
same size as in the training from scratch experiments.

\begin{figure}[t]
  \centering
  \includegraphics[width=\textwidth,height=5.5cm]{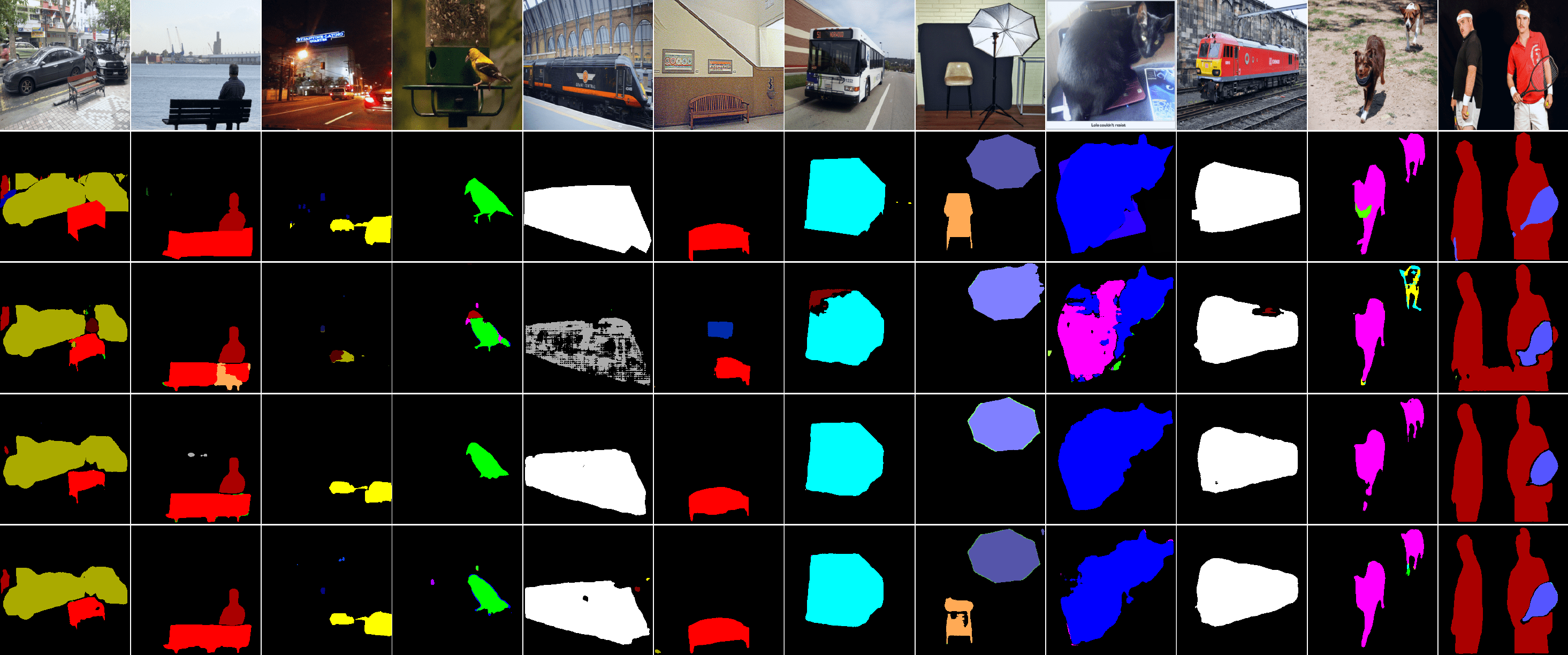}
  \caption{
    Sample segmentation results on the MSCOCO 2015 detection dataset using
    training from scratch models. The
    first and second rows are the original images and the corresponding ground
    truth, respectively. The third, fourth, and fifth rows are the
    segmentation results of models using deconvolutional layers, iPixelDCL,
    and PixelDCL, respectively.
  }
  \label{fig:coco_result}
\end{figure}

\textbf{Analysis of Results:}
Some sample segmentation results of U-Net using
deconvolutional layers (DCL), iPixelDCL, and PixelDCL on the PASCAL
2012 segmentation dataset and the MSCOCO 2015 detection dataset are
given in Figures~\ref{fig:pascal_result} and~\ref{fig:coco_result},
respectively. We can see that U-Net models using iPixelDCL and
PixelDCL can better capture the local information of images than the
same base model using regular deconvolutional layers. By using pixel
deconvolutional layers, more spacial features such as edges and
shapes are considered when predicting the labels of adjacent pixels.

\begin{table}[t]
\centering \caption{
Semantic segmentation results on the PASCAL 2012
segmentation dataset and MSCOCO 2015 detection dataset. We compare
the same base U-Net model and fine-tuned DeepLab-ResNet using three different
up-sampling methods in decoders; namely regular deconvolution layer
(DCL), the proposed input pixel deconvolutional layer (iPixelDCL)
and pixel deconvolutional layer (PixelDCL).
The pixel accuracy and mean IOU are used as performance measures.
} \label{table:comp1}

\begin{tabular}{ l| l | c | c }
    \hline
    \textbf{Dataset} &\textbf{Model}      & \textbf{Pixel Accuracy} & \textbf{Mean IOU} \\
    \hline
    \multirow{3}{*}{PASCAL 2012}& U-Net + DCL            & 0.816161  & 0.415178 \\
    &U-Net + iPixelDCL            & 0.817129  & 0.448817 \\ 
    &U-Net + PixelDCL & \textbf{0.822591}    & \textbf{0.455972} \\
    \hline
    \multirow{3}{*}{MSCOCO 2015}
    &U-Net + DCL            & 0.809327           & 0.349769 \\ 
    &U-Net + iPixelDCL      & 0.809239           & 0.360216 \\ 
    &U-Net + PixelDCL       & \textbf{0.811575}  & \textbf{0.371805} \\
    \hline
    \multirow{3}{*}{PASCAL 2012}
    &DeepLab-ResNet + DCL            & 0.929562           & 0.727036 \\ 
    &DeepLab-ResNet + iPixelDCL      & \textbf{0.934493}  & \textbf{0.738552} \\ 
    &DeepLab-ResNet + PixelDCL       & 0.931287           & 0.735585 \\\hline
\end{tabular}
\end{table}

Moreover, the semantic segmentation results demonstrate that the
proposed models tend to produce smoother outputs than the model
using deconvolution. We also observe that, when the training epoch
is small (e.g., 50 epochs), the model that employs PixelDCL has
better segmentation outputs than the model using iPixelDCL. When the
training epoch is large enough (e.g., 100 epochs), they have similar
performance, though PixelDCL still outperforms iPixelDCL in most
cases. This indicates that PixelDCL is more efficient and effective,
since it has much fewer parameters to learn.

Table~\ref{table:comp1} shows the evaluation results in terms of
pixel accuracy and mean IOU on the two datasets. The U-Net models
using iPixelDCL and PixelDCL yield better performance than the same
base model using regular deconvolution. The model using PixelDCL
slightly outperforms the model using iPixelDCL. For the models
fine-tuned from Deeplab-ResNet, the models using iPixelDCL and
PixelDCL have better performance than the model using DCL, with
iPixelDCL performs the best. In semantic segmentation, mean IOU is a
more accuracy evaluation measure than pixel
accuracy~\citep{everingham2010pascal}. The models using pixel
deconvolution have better evaluation results on mean IOU than the
base model using deconvolution.

\subsection{Image Generation}

\textbf{Experimental Setup:}
The dataset used for image generation is the celebFaces attributes
(CelebA) dataset~\citep{liu2015faceattributes}. To avoid the
influence of background, the images have been preprocessed so that
only facial information is retained. The image generation task is to
reconstruct the faces excluding backgrounds in training images. The
size of images is $64\times64\times3$. We use the standard
variational auto-encoder (VAE)~\citep{kingma2014stochastic} as our base
model for image generation. The decoder part in standard VAE employs
deconvolutional layers for up-sampling. We apply our proposed
PixelDCL to replace deconvolutional layers in decoder while keeping
all other components the same. The kernel size in DCL is 6$\times$6,
which has more parameters than PixelDCL with
2 sets of 3$\times$3 and 1 set of 2$\times$2 kernels.

\begin{figure}[t]
  \centering
  \includegraphics[width=\textwidth]{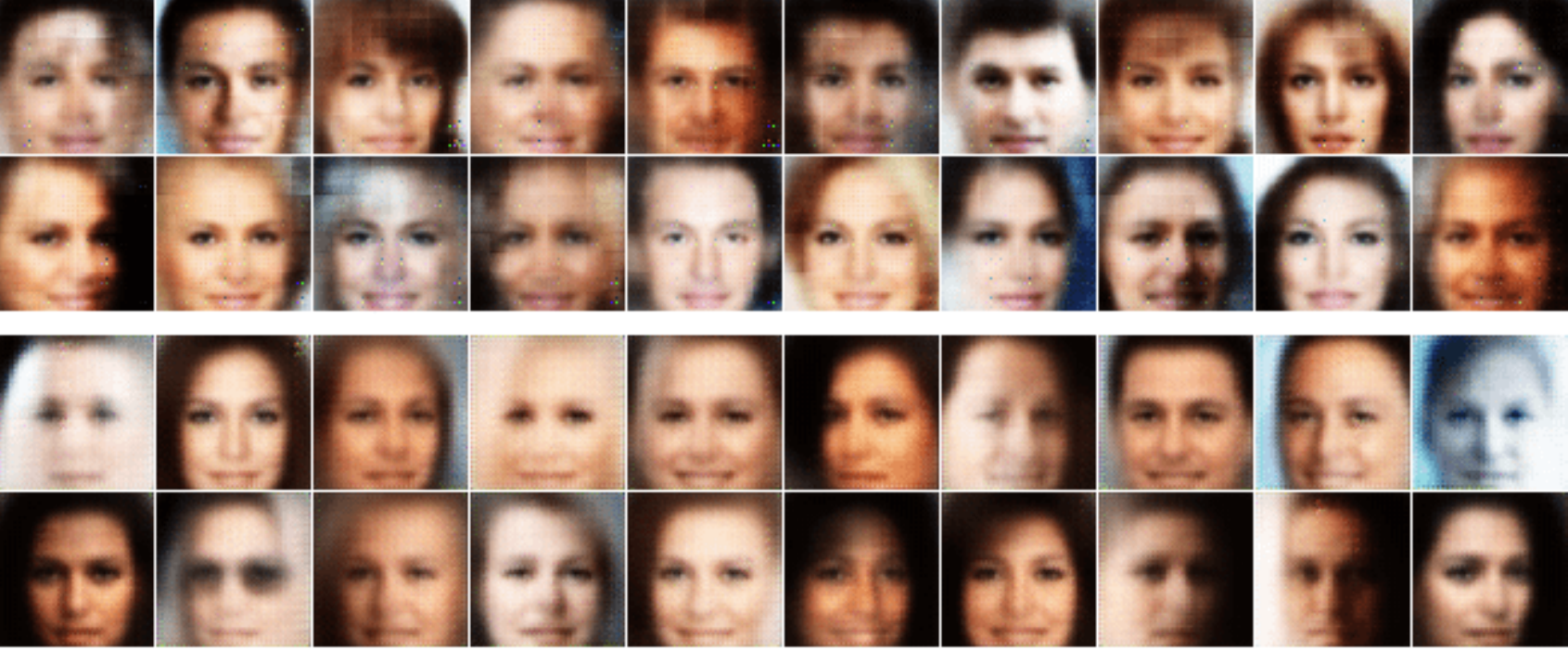}\vspace{-0.3cm}
  \caption{
    Sample face images generated by VAEs when trained on the CelebA dataset.
    The first two rows are images generated by a standard VAE with
    deconvolutional layers for up-sampling. The last two rows
    generated by the same VAE model, but using PixelDCL for up-sampling.
  }
  \label{fig:vae_result}
\end{figure}

\textbf{Analysis of Results:}
Figure~\ref{fig:vae_result} shows the generated faces using VAEs
with regular deconvolution (baseline) and PixelDCL in decoders. Some
images generated by the baseline model suffer from apparent
checkerboard artifacts, while none is found on the images generated
by the model with PixelDCL. This demonstrates that the proposed
pixel deconvolutional layers are able to establish direct
relationships among adjacent pixels on generated feature maps and
images, thereby effectively overcoming the checkerboard problem. Our
results demonstrate that PixelDCL is very useful for generative
models since it can consider local spatial information and produce
photo-realistic images without the checkerboard problem.

\subsection{Timing Comparison}

\begin{table}[t]
\centering
  \caption{
    Training and prediction time on semantic segmentation using the PASCAL 2012
    segmentation dataset on a Tesla K40 GPU.
    We compare the training time of 10 epochs and prediction time of 2109 images
    for the same base U-Net model using three different methods for
    up-sampling in the decoders; namely DCL, iPixelDCL, and PixelDCL.
  } \label{table:comp2}
\begin{tabular}{  l | c | c }
   \hline
   \textbf{Model}      & \textbf{Training time} & \textbf{Prediction time} \\
   \hline
   U-Net + DCL         &  365m26s   & 2m42s \\ 
   U-Net + iPixelDCL   & 511m19s    & 4m13s \\
   U-Net + PixelDCL    & 464m31s    & 3m27s \\
   \hline
\end{tabular}
\end{table}

Table~\ref{table:comp2} shows the comparison of the training and
prediction time of the U-Net models using DCL, iPixelDCL, and
PixelDCL for up-sampling. We can see that the U-Net models using
iPixelDCL and PixelDCL take slightly more time during training and
prediction than the model using DCL, since the intermediate feature
maps are generated sequentially. The model using PixelDCL is more
efficient due to reduced dependencies and efficient implementation
discussed in Section~\ref{sec:pixelNet}. Overall, the increase in
training and prediction time is not dramatic, and thus we do not
expect this to be a major bottleneck of the proposed methods.

\section{Conclusion}\label{conclusion}

In this work, we propose pixel deconvolutional layers that can solve
the checkerboard problem in deconvolutional layers. The checkerboard
problem is caused by the fact that there is no direct relationship
among intermediate feature maps generated in deconvolutional layers.
PixelDCL proposed here try to add direct dependencies among these
generated intermediate feature maps. PixelDCL generates intermediate
feature maps sequentially so that the intermediate feature maps
generated in a later stage are required to depend on previously
generated ones. The establishment of dependencies in PixelDCL can
ensure adjacent pixels on output feature maps are directly related.
Experimental results on semantic segmentation and image generation
tasks show that PixelDCL is effective in overcoming the checkerboard
artifacts. Results on semantic segmentation also show that PixelDCL
is able to consider local spatial features such as edges and shapes,
leading to better segmentation results.
In the future, we
plan to employ our PixelDCL in a broader class of models, such as
the generative adversarial networks (GANs).

\bibliography{nips}
\bibliographystyle{iclr2018_conference}

\end{document}